\documentclass{article} 
\usepackage[final]{colm2026_conference}

\usepackage{microtype}
\usepackage{hyperref}
\usepackage{url}
\usepackage{booktabs}
\usepackage{graphicx}

\usepackage{lineno}

\definecolor{darkblue}{rgb}{0, 0, 0.5}
\hypersetup{colorlinks=true, citecolor=darkblue, linkcolor=darkblue, urlcolor=darkblue}

\title{MS-GLA: Multi-Scale Gated Linear Attention for Addressing Representational Bottlenecks via Multi-Temporal Resolution}

\author{Prasoon Dev, Anirudh Sankar, Vasudeva Varma\\
Language Technologies Research Center\\
International Institute of Information Technology Hyderabad\\
Hyderabad, Telangana, India \\
\texttt{\{prasoon.dev,anirudh.sankar\}@research.iiit.ac.in, vv@iiit.ac.in} \\
}

\usepackage{amsmath,amsfonts,bm}

\def\eqref#1{equation~\ref{#1}}

\def\1{\bm{1}}

\def\rk{{\textnormal{k}}}

\def\rv{{\textnormal{v}}}

\def\vk{{\bm{k}}}

\def\vo{{\bm{o}}}

\def\vq{{\bm{q}}}

\def\vv{{\bm{v}}}

\def\mG{{\bm{G}}}

\def\mK{{\bm{K}}}

\def\mM{{\bm{M}}}

\def\mO{{\bm{O}}}

\def\mQ{{\bm{Q}}}

\def\mS{{\bm{S}}}

\def\mV{{\bm{V}}}

\DeclareMathAlphabet{\mathsfit}{\encodingdefault}{\sfdefault}{m}{sl}
\SetMathAlphabet{\mathsfit}{bold}{\encodingdefault}{\sfdefault}{bx}{n}

\def\sR{{\mathbb{R}}}

\begin{document}

\ifcolmsubmission
\linenumbers
\fi

\maketitle

\begin{abstract}
Gated Linear Attention (GLA) Transformers advance linear recurrent models through data-dependent gating, but face a core limitation: the fixed-capacity memory matrices across all heads operate at a single temporal resolution, where each token is processed individually, forcing them to simultaneously encode local syntactic patterns and long-range semantic structure, creating a representational bottleneck that gating alone is insufficient to resolve. We introduce Multi-Scale Gated Linear Attention (MS-GLA), which addresses this by distributing attention heads across multiple temporal resolutions. Coarser resolutions pool longer token spans naturally specializing toward long-range dependencies, while finer head groups retain sensitivity to local syntactic structure. A learnable, input-dependent fusion layer dynamically recombines head group outputs at each timestep, expanding effective memory capacity without increasing per-head state size. This multi-resolution decomposition draws on principles from Multi-Scale State-Space Models (MS-SSM), adapting them to the gated linear attention setting. We evaluate MS-GLA on language modeling, recall-intensive tasks, and long-context generalization. Across all settings, MS-GLA consistently achieves higher accuracy and lower 
perplexity than GLA at matched parameter counts, with up to 18.9\% improvement on recall-intensive tasks and 9.5\% lower average perplexity on language modeling benchmarks, validating multi-temporal resolution decomposition as a principled and effective extension of Gated Linear Attention.
\end{abstract}

\section{Introduction}
Transformers \citep{Vaswani+2017} have long been the gold standard for sequence modeling, their reliance on softmax attention yields quadratic computational complexity with respect to sequence length. To overcome this, the field has been rapidly shifting towards sub-quadratic, linear-time models. Architectures like Linear Recurrent Neural Networks (RNNs), State Space Models (SSMs) like Mamba \citep{gu2024mambalineartimesequencemodeling}, and Linear Attention variants (like RetNet \citep{sun2024retentive}) treat sequence mixing as a linear recurrence. This is highly efficient, allowing for an $O(1)$ memory footprint during inference and parallelized training. Recently, Gated Linear Attention (GLA) \citep{yang2024gla} took this a step further. By adding data-dependent gating, GLA gave models the ability to actually choose what context to remember and what obsolete information to forget, effectively bridging the performance gap with standard Transformers while keeping training highly hardware-efficient.

While data-dependent gating improves how information is retained over time, GLA still operates at a single temporal resolution, where its fixed-capacity memory matrix must summarize the entire sequence history. This creates an inherent representational bottleneck, especially for language, which naturally exhibits a multi-scale structure \citep{tamkin2020languageprismspectralapproach, nawrot2022hierarchicaltransformersefficientlanguage}: a single shared state must simultaneously capture rapidly shifting, high-frequency syntactic details alongside slowly evolving, low-frequency semantic structures. Expressive gates might allow a model to choose what to remember or forget, but they cannot fix the underlying problem of how to stretch a limited memory budget across these wildly different timescales. 

To address this, we propose Multi-Scale Gated Linear Attention (MS-GLA). Instead of forcing the model to operate at a single temporal resolution handling all timescales, MS-GLA explicitly breaks the input sequence down into multiple temporal resolutions and assigns dedicated groups of GLA heads to each. By operating on temporally pooled tokens, the coarser head groups naturally expand their receptive fields, allowing their gates to specialize entirely in tracking long-range semantic dependencies. Meanwhile, the finer head groups process tokens at their original resolution, retaining strict sensitivity to local syntactic structure. This multi-resolution decomposition is grounded in the principles of Multi-Scale State-Space Models (MS-SSM) \citep{karami2025msssm}, which demonstrate that assigning recurrent branches to distinct temporal scales yields complementary specialization that a single shared state cannot achieve. MS-GLA adapts this insight to the gated linear attention setting, replacing SSM branches with groups of GLA heads operating on pooled token sequences. 

Crucially, MS-GLA achieves this division of labor without bearing the overhead of typical ensemble approaches. We aren't instantiating full-sized models at every scale or inflating the parameter count. Instead, MS-GLA simply redistributes the baseline model's existing budget of attention heads across these different time scales (for example, assigning two heads to fine resolutions and two heads to coarser scales). The outputs from these specialized groups are then upsampled via a causal zero-order hold and dynamically recombined at each timestep using a learnable fusion layer. This allows the model to adaptively weigh fine-grained details and coarse-grained information based on the immediate context. Furthermore, MS-GLA employs non-overlapping causal average pooling, ensuring full compatibility with the hardware-efficient chunkwise training algorithms that make GLA practical at scale.

\section{Background and Related Work}

\subsection{Linear Attention and Gated Linear Attention (GLA)}
Transformers use softmax attention which do not scale well to long sequences because the complexity is quadratic with respect to sequence length. This has motivated the development of Linear Attention mechanisms which achieve linear-time complexity with respect to sequence length \citep{10.5555/3524938.3525416}. To achieve linear complexity in sequence length, the softmax operation is replaced with a kernel with an associated feature map \citep{10.5555/3524938.3525416}. This enables the reformulation of attention as a linear recurrent neural network (RNN). Prior work has shown that an unnormalized linear kernel works best \citep{sun2024retentive}. Therefore, the recurrent update for the hidden state \(\mS_t \in \sR^{d_k \times d_v}\) and the output \(o_t\) at time step $t$ is given by:
\begin{equation}
    \mS_t = \mS_{t-1} + \rk^T_t \rv_t
\end{equation}
\begin{equation}
    \vo_t = \vq_t \mS_t
\end{equation}
where $\vq_t$, $\vk_t$ and $\vv_t$ denote the query, key and value vectors respectively.

The \textit{chunkwise parallel form} \citep{hua2022transformerqualitylineartime} of linear attention enables partially parallel training with subquadratic complexity. Striking a balance between the computationally slow \textit{recurrent} form and the computationally expensive \textit{parallel} form of attention. In the \textit{chunkwise parallel} form, the input sequence is divided into non-overlapping chunks of size \(C\). The recurrence updates are thus split into two phases: an inter-chunk recurrence and an intra-chunk computation. The inter-chunk recurrence works sequentially to update the chunk level hidden state and the intra-chunk output computation computes the outputs in parallel. That is formally, the updates for chunk \(i\) are given by:
\begin{equation}
    \mS[i+1] = \mS[i] + \mK[i]^T \mV[i]
\end{equation}
\begin{equation}
    \mO[i+1] = \mQ[i+1]\mS[i] + \left((\mQ[i+1]\mK[i+1]^T) \odot \mM\right)\mV[i+1]
\end{equation}
where $\mK[i], \mV[i], \mQ[i+1]$ are the stacked token vectors for their respective chunks, and $\mM$ is the causal mask.

While computationally efficient, standard linear attention lacks a decay term, which makes it difficult for the model to "forget" historically irrelevant information \citep{buckman2024}. This has been hypothesized to potentially limit performance of  on long-context tasks. Gated Linear Attention (GLA) resolves this by incorporating a data-dependent 2D forget gate \citep{yang2024gla} $\mG_t \in (0,1)^{d_k \times d_v}$ into the linear recurrence. The generalized gated update is expressed as:

\begin{equation}
    \mS_t = \mG_t \odot \mS_{t-1} + \vk_t^T \vv_t
\end{equation}

where $\odot$ denotes the Hadamard product. To maintain a balance between parameter efficiency, state size, and hardware-efficient training, GLA specifically parametrizes the forget gate as $G_t = \alpha_t^T \mathbf{1}$, which simplifies the hidden state update to a row-wise scaling:

\begin{equation}
    \mS_t = \text{Diag}(\alpha_t)\mS_{t-1} + \vk_t^T \vv_t
\end{equation}

where the data-dependent gating vector $\alpha_t \in \mathbb{R}^{1 \times d_k}$ is generated dynamically at each time step using a low-rank linear projection of the input $x_t$ followed by a sigmoid activation. This data-dependent gating mechanism significantly enhances the expressive power and length extrapolation capabilities of the model while retaining full compatibility with the sub-quadratic chunkwise parallel training form.

\subsection{Mixture-of-Memories (MoM)}
A related line of work, Mixture-of-Memories (MoM) \citep{du2025momlinearsequencemodeling}, also targets the fixed-capacity memory bottleneck of linear sequence models, but through a fundamentally different mechanism. MoM maintains several independent memory states and uses a learned top-$k$ router to direct individual \emph{tokens} to a sparse subset of these memories at each step, reducing interference by ensuring that only a subset of memory states is updated by any given token. MS-GLA instead partitions the \emph{temporal resolution} at which the sequence itself is processed: deterministic, parameter-free average pooling produces multiple coarser views of the same sequence, with dedicated GLA head groups assigned to each view.
Where MoM asks \emph{which memory should this token update}, MS-GLA asks \emph{at what temporal resolution should this sequence be processed}. The two mechanisms are complementary rather than competing, and we
view combining MoM-style token routing with MS-GLA's multi-resolution branches as a promising direction for future work.

\subsection{Multi-Scale Sequence Modeling and MS-SSM}
Real-world signals such as text, images, and time series naturally exhibit multi-scale
structure, with patterns ranging from fine-grained, high-frequency details to coarse,
global trends \citep{article,shi2023sequencemodelingmultiresolutionconvolutional}. Multi-Resolution
Analysis (MRA) techniques such as the Stationary Wavelet Transform \citep{nason1995} have
long been used to capture such hierarchies. Multi-Scale State-Space Models (MS-SSM)
\citep{karami2025msssm} adapt this MRA framework to deep sequence modeling by replacing
fixed wavelet bases with trainable, causal depthwise 1D convolutions, recursively applying
dilated convolutions to decompose the input sequence:
\begin{equation}
    [a_s; d_s] = \text{Conv1d}(1, 2, L, 2^{s-1})[a_{s-1}]
\end{equation}
This transforms the sequence into a multi-resolution representation $x_t \mapsto \hat{x}_t
\in \mathbb{R}^{S+1}$, where higher scales capture coarse-grained structure over larger
receptive fields and lower scales retain sharp, localized detail. Each scale is fed into a
separate State Space Model (SSM) operating in parallel. Since effective memory capacity in
linear recurrent models is inversely proportional to the distance of the state transition
matrix's eigenvalues from the unit circle \citep{agarwal2024spectralstatespacemodels},
MS-SSM initializes coarser branches with transition matrix values closer to 1 (slower
forgetting, favoring long-range dependencies) and finer branches with smaller values
(faster forgetting, favoring local dynamics). The branch outputs are then combined via an
input-dependent scale-mixer, $z_t = \text{LinearE}(x_t)\, y_t$ \citep{karami2025msssm},
which dynamically routes information across scales based on the raw input token.

\section{Multi-Scale Gated Linear Attention (MS-GLA)}
To alleviate the limitations of modeling an entire sequence through a single temporal resolution, where each recurrent update operates on individual tokens and must simultaneously encode both local syntactic patterns and long-range semantic structure, we introduce Multi-Scale Gated Linear Attention (MS-GLA). Instead of relying on one recurrent pathway to represent both rapidly varying local patterns and slower long-range structure, MS-GLA processes the input through multiple GLA heads operating at different temporal scales. This provides an architectural bias toward modeling information at multiple timescales within the same layer \citep{karami2025msssm, buckman2024}.

\subsection{Multi-Scale Decomposition via Temporal Pooling}
Given input hidden states $\mathbf{X} \in \mathbb{R}^{L \times d}$, where $L$ is the sequence length and $d$ is the hidden dimension, let $\mathcal{S}$ denote a set of predefined temporal scales (e.g.\ $\mathcal{S}=\{1,2,4\}$). For each scale $s \in \mathcal{S}$, MS-GLA constructs a lower-resolution sequence by applying non-overlapping average pooling over blocks of size $s$ \citep{karami2025msssm, nawrot2022hierarchicaltransformersefficientlanguage}. Throughout this work, we use temporal scale and temporal resolution interchangeably to refer to the pooling factor $s$: a larger $s$ produces a coarser, lower-frequency representation in which every s consecutive input positions are pooled into a single token, whereas $s = 1$ preserves the original token-level resolution.

When the sequence length $L$ is not divisible by $s$, the sequence is right-padded with zeros before pooling. If an attention mask $\mathbf{M} \in \{0,1\}^L$ is provided, pooling is performed in a masked manner. We denote by $\mathbf{X}^{(s)}$ the pooled hidden-state sequence at scale $s$. The pooled representation for the $i$-th block is
\begin{equation}
\mathbf{X}^{(s)}_i
=
\frac{\sum_{j=0}^{s-1} \mathbf{X}_{i s + j}\,\mathbf{M}_{i s + j}}
{\max\!\left(1,\sum_{j=0}^{s-1}\mathbf{M}_{i s + j}\right)}.
\end{equation}
The numerator sums the hidden states of the $s$ tokens in the $i$-th block, 
weighted by their mask values so that padding tokens contribute nothing. 
The denominator normalizes by the number of valid (unmasked) tokens in the 
block, with the $\max(1, \cdot)$ guard preventing division by zero in 
fully-padded blocks.

In the absence of masking, this reduces to standard average pooling. After padding, the pooled sequence length is approximately $\lceil L/s \rceil$. As $s$ increases, each recurrent update summarizes a larger temporal block, biasing coarser resolutions toward longer-range dependencies.

\subsection{Scale-Specific GLA Heads}
MS-GLA processes each temporal scale through dedicated GLA heads, while
keeping the total attention-head and recurrent-state budget identical to the
baseline. Each branch is an independent GLA module that takes its
scale-specific pooled sequence $\mathbf{X}^{(s)}$ as input. Branches share
the same architectural form as GLA: query, key, value projections,
data-dependent gating, and normalization, but maintain entirely separate
learned parameters. This means that each branch's gates and projections can
specialize freely to the temporal scale it operates at, without any parameter
sharing across scales.

Rather than adding new heads, MS-GLA partitions the baseline model's
$H$ attention heads across the set of scales $\mathcal{S}$, assigning $h_s$
heads to the branch at scale $s$, such that
\begin{equation}
\sum_{s \in \mathcal{S}} h_s = H.
\end{equation}
For example, with $H=4$ heads and scales $\mathcal{S}=\{1,2,4\}$, the head
allocation $\mathbf{h}=[2,1,1]$ assigns two heads to the finest scale ($s=1$)
and one head each to the coarser scales ($s=2$ and $s=4$). The key and value
projection dimensions are split proportionally, so each branch receives a
fraction $h_s / H$ of the total key-value budget.

\subsection{Causal Upsampling and Learnable Scale Fusion}
After branch-wise sequence mixing, all branch outputs must be mapped back to the original token resolution. For a branch operating at scale $s$, the branch output is repeated $s$ times along the sequence dimension and then shifted to the right by $s-1$ positions, with zeros inserted at the beginning. This implements a causal hold mechanism: the output corresponding to a pooled block is not revealed until all tokens in that block have been observed \citep{karami2025msssm}.

We denote the causally aligned output of branch $s$ after upsampling by $\hat{\mathbf{Y}}^{(s)} \in \mathbb{R}^{L \times d}$. MS-GLA then combines the aligned branch outputs using a learnable, input-dependent fusion module \citep{karami2025msssm}. For each timestep $t$, the routing weights $\mathbf{w}_t$ are computed from the unpooled hidden state $\mathbf{x}_t$ using a learned fusion projection with weight matrix $\mathbf{W}_{\mathrm{fuse}} \in \mathbb{R}^{|\mathcal{S}| \times d}$ and bias vector $\mathbf{b}_{\mathrm{fuse}} \in \mathbb{R}^{|\mathcal{S}|}$:
\begin{equation}
\mathbf{w}_t = \mathrm{Softmax}(\mathbf{W}_{\mathrm{fuse}}\mathbf{x}_t + \mathbf{b}_{\mathrm{fuse}}).
\end{equation}

The final layer output at timestep $t$, denoted by $\mathbf{O}_t$, is
\begin{equation}
\mathbf{O}_t = \sum_{s \in \mathcal{S}} w_{t,s}\,\hat{\mathbf{Y}}^{(s)}_t.
\end{equation}

\subsection{Compatibility with Chunkwise GLA Training}
\label{sec:chunkwise}
A key practical advantage of standard GLA is its hardware-efficient chunkwise parallel training formulation. MS-GLA preserves compatibility with this training regime \citep{yang2024gla, hua2022transformerqualitylineartime}. Since the multi-scale decomposition uses static, non-overlapping pooling, each branch still operates on a contiguous sequence. As a result, every scale-specific branch can apply the same chunkwise GLA computations as the baseline model, but on a shorter sequence whose length decreases with the branch scale.

Consequently, MS-GLA extends GLA with multi-resolution sequence modeling while retaining the underlying chunkwise training structure of the original architecture. Although adding multiple branches introduces additional computation, coarser branches operate on proportionally shorter sequences, which mitigates part of this overhead and helps maintain practical training efficiency.

\section{Experimental Setup}
Our main experiments study whether Multi-Scale Gated Linear Attention (MS-GLA) 
improves over standard Gated Linear Attention (GLA) under matched model and 
training budgets. We focus on autoregressive language modeling and long-context 
evaluation, and compare MS-GLA variants against a GLA baseline, trained using 
the \citet{yang2024gla}'s implementation in \verb|flash-linear-attention| 
\citep{yang2024fla} with the \verb|flame| training framework \citep{yang2025flame} 
under an identical architecture and optimization recipe, isolating the effect 
of multi-scale temporal decomposition from changes in model capacity or training 
procedure.

\subsection{Model variants.}
All models share the same decoder-only backbone and differ only in the sequence-mixing module. The common backbone uses 24 layers, hidden size 1024, and 4 total attention heads. This choice of 4 heads is inherited directly from the original GLA configuration: \citet{yang2024gla} demonstrate in their ablation study that 4 heads represents a favorable tradeoff between training throughput and perplexity, with larger head counts degrading perplexity and smaller head counts offering only marginal gains at the cost of significantly higher memory usage. We adopt this finding unchanged, as our contribution lies in how the fixed head budget is allocated across temporal scales, not in the total number of heads.

While maintaining the same parameter count as the baseline GLA model, we train and evaluate five multiscale variants of MS-GLA:
\[
\{1,2\}, \quad \{1,4\}, \quad \{2,4\}, \quad \{1,2,4\}, \quad \{1,2,4,8\}.
\]

For MS-GLA, the total head budget is partitioned across scales while keeping the overall attention/state budget fixed. The corresponding head allocations are:
\[
[2,2], \quad [2,2], \quad [2,2], \quad [2,1,1], \quad [1,1,1,1].
\]

\subsection{Training details.}
\label{train_deets}
All models are trained from scratch on the FineWeb-Edu corpus \citep{lozhkov2024fineweb-edu}
with the same tokenizer using AdamW \citep{article-adam}. We train 340M parameter models for
7B tokens, following the compute-optimal token-to-parameter ratio prescribed by the
Chinchilla scaling laws \citep{hoffmann2022trainingcomputeoptimallargelanguage} for a model
of this scale. All reported comparisons are made at matched parameter count and matched
training tokens under an identical optimization recipe. Full training configuration --
including optimizer hyperparameters, batch size, warmup schedule, hardware, precision, and
random seed -- along with our released code, is provided in Appendix~\ref{app:training_details}.

\section{Results}
To assess whether multi-temporal resolution decomposition yields broad and robust improvements, we evaluate five MS-GLA configurations ($\{1,2\}$, $\{1,4\}$, $\{2,4\}$,
$\{1,2,4\}$, and $\{1,2,4,8\}$) against a GLA baseline across three axes of model quality, followed by two analyses that probe how these gains are achieved. First, \textbf{language modeling} measures general sequence understanding, capturing both local syntax and long-range semantics via perplexity on Wikitext \citep{merity2017pointer} and LAMBADA
\citep{paperno-EtAl:2016:P16-1}, along with zero-shot accuracy on LAMBADA, PIQA \citep{Bisk2020}, HellaSwag \citep{zellers2019hellaswag}, and Winogrande \citep{ai2:winogrande}. Second, \textbf{recall-intensive tasks} evaluate the model's ability to retrieve precise in-context information without memory saturation, using F1 score on SWDE \citep{lockard-etal-2019-openceres}, FDA \citep{arora2025languagemodelsenablesimple}, and SQuAD \citep{Rajpurkar2018SQuAD2} as modified in \citet{arora2024simple}. Third, \textbf{long-context generalization} probes robustness far beyond the training horizon by measuring perplexity as a function of token
position on SlimPajama \citep{cerebras2023slimpajama} and PG19 \citep{Rae2020Compressive} at sequence lengths up to $15\times$ the training context (30,720 tokens), where flat curves indicate successful extrapolation and divergence signals recurrent state saturation. We then examine \textbf{fusion layer routing stability} to verify that this
long-context robustness reflects genuine multi-scale usage rather than collapse onto a single branch, and finally quantify \textbf{computational efficiency} to confirm that any gains are achieved without disproportionate cost in compute, memory, or wall-clock training time.

\subsection{Language Modeling}
\begin{table}[t]
\begin{center}
\resizebox{\textwidth}{!}{%
\begin{tabular}{llcccccccc}
\toprule
\multicolumn{1}{c}{\bf Scale} & \multicolumn{1}{c}{\bf Model} & \multicolumn{1}{c}{\bf Wiki.\ ppl$\downarrow$} & \multicolumn{1}{c}{\bf LMB.\ ppl$\downarrow$} & \multicolumn{1}{c}{\bf LMB.\ acc$\uparrow$} & \multicolumn{1}{c}{\bf PIQA$\uparrow$} & \multicolumn{1}{c}{\bf Hella.$\uparrow$} & \multicolumn{1}{c}{\bf Wino.$\uparrow$} & \multicolumn{1}{c}{\bf Avg.\ score$\uparrow$} & \multicolumn{1}{c}{\bf Avg.\ ppl$\downarrow$} \\
\midrule
\textit{340M}   & GLA                  & 19.63          & 22.26          & 19.33          & 64.20          & 33.94          & 50.28          & 41.94          & 20.94 \\
\textit{7B tok} & MS-GLA $\{1,2\}$     & 19.25          & 20.99          & 19.68          & 65.18          & 34.89          & 49.17          & 42.23          & 20.12 \\
                & MS-GLA $\{1,4\}$     & \textbf{17.43} & 21.69          & 19.89          & 63.87          & \textbf{35.49} & 51.07          & 42.58          & 19.56 \\
                & MS-GLA $\{2,4\}$     & 21.66          & 31.47          & 16.15          & 65.18          & 34.17          & \textbf{52.57} & 42.01          & 26.56 \\
                & MS-GLA $\{1,2,4\}$   & 18.33          & \textbf{19.59} & \textbf{20.69} & \textbf{65.83} & 35.01          & 49.88          & \textbf{42.85} & \textbf{18.96} \\
                & MS-GLA $\{1,2,4,8\}$ & 18.80          & 20.18          & 20.34          & 65.34          & 34.53          & 49.01          & 42.31          & 19.49 \\
\bottomrule
\end{tabular}%
}
\end{center}
\caption{Language modeling perplexity and zero-shot downstream task performance of GLA and MS-GLA variants. Best results in each column are shown in \textbf{bold}.}

\label{tab:main-results-7b}
\end{table}

Table~\ref{tab:main-results-7b} reports language modeling and zero-shot downstream results across all six models. The clearest pattern is that including the finest temporal resolution ($s=1$) is necessary for reliable improvement: every configuration that retains at least one native-resolution branch outperforms GLA in aggregate, while the purely coarse-grained $\{2,4\}$ configuration degrades substantially across most metrics.

MS-GLA $\{1,2,4\}$ achieves the best overall performance, reducing average perplexity by roughly 9.5\% and improving mean downstream accuracy across Lambada, PIQA, and HellaSwag relative to the GLA baseline. MS-GLA $\{1,4\}$ leads on Wikitext perplexity and HellaSwag but lags behind $\{1,2,4\}$ on average, suggesting that skipping the intermediate scale creates a coherence gap despite strong structural modeling. The four-scale configuration $\{1,2,4,8\}$ remains competitive overall but falls slightly short of $\{1,2,4\}$ across perplexity and accuracy, consistent with head-budget dilution as representational capacity is spread across an additional scale. Winogrande is the one exception where fine-resolution configurations offer no clear advantage, and the $\{2,4\}$ variant, which entirely omits the fine-grained scale ($s=1$), records the highest individual score on this benchmark, a pattern we discuss further in Section~\ref{sec:sc vs bc}.

\subsection{Recall-Intensive Tasks}
\begin{table}[t]
\begin{center}
\begin{tabular}{lcccc}
\toprule
\multicolumn{1}{c}{\bf Model} & \multicolumn{1}{c}{\bf SWDE F1$\uparrow$} & \multicolumn{1}{c}{\bf FDA F1$\uparrow$} & \multicolumn{1}{c}{\bf SQuAD F1$\uparrow$} & \multicolumn{1}{c}{\bf Avg.\ F1$\uparrow$} \\
\midrule
GLA                   & 7.86          & 5.52          & 8.84           & 7.41 \\
MS-GLA $\{1,2\}$      & 9.26          & 5.95          & 9.79           & 8.33 \\
MS-GLA $\{1,4\}$      & 8.28          & \textbf{6.06} & 10.32          & 8.22 \\
MS-GLA $\{2,4\}$      & 5.96          & 5.42          & 8.21           & 6.53 \\
MS-GLA $\{1,2,4\}$    & \textbf{9.30} & 5.98          & \textbf{11.15} & \textbf{8.81} \\
MS-GLA $\{1,2,4,8\}$  & 7.59          & 5.27          & 11.05          & 7.97 \\
\bottomrule
\end{tabular}
\end{center}
\caption{Performance comparison of GLA and various MS-GLA configurations on recall-intensive benchmarks. F1 scores are reported for each task, with the highest values highlighted in \textbf{bold}.}
\label{tab:recall-results}
\end{table}

Table~\ref{tab:recall-results} reports recall-intensive generation performance. The ranking of configurations mirrors the language modeling results closely. MS-GLA $\{1,2,4\}$ achieves the highest average F1, representing roughly an 18.9\% improvement over the GLA baseline, with leading scores on both SWDE and SQuAD. MS-GLA $\{1,2\}$ and $\{1,4\}$ also improve over GLA in aggregate, with the latter achieving the best FDA score among all models. MS-GLA $\{1,2,4,8\}$ outperforms the baseline overall but falls noticeably behind $\{1,2,4\}$, particularly on SWDE where the gap is most pronounced.

The strictly coarse-scale configuration $\{2,4\}$ again performs worst, dropping below the GLA baseline on SWDE and SQuAD and recording the lowest average F1 among all MS-GLA variants. The consistency of this failure across both language modeling and recall tasks confirms that local token-level processing is not merely complementary but essential i.e. coarser branches alone are insufficient for reliable in-context retrieval.

\subsection{Long-Context Generalization}
\label{sec:LongContextGeneralization}
\begin{figure}[t]
\begin{center} 
\includegraphics[width=\linewidth]{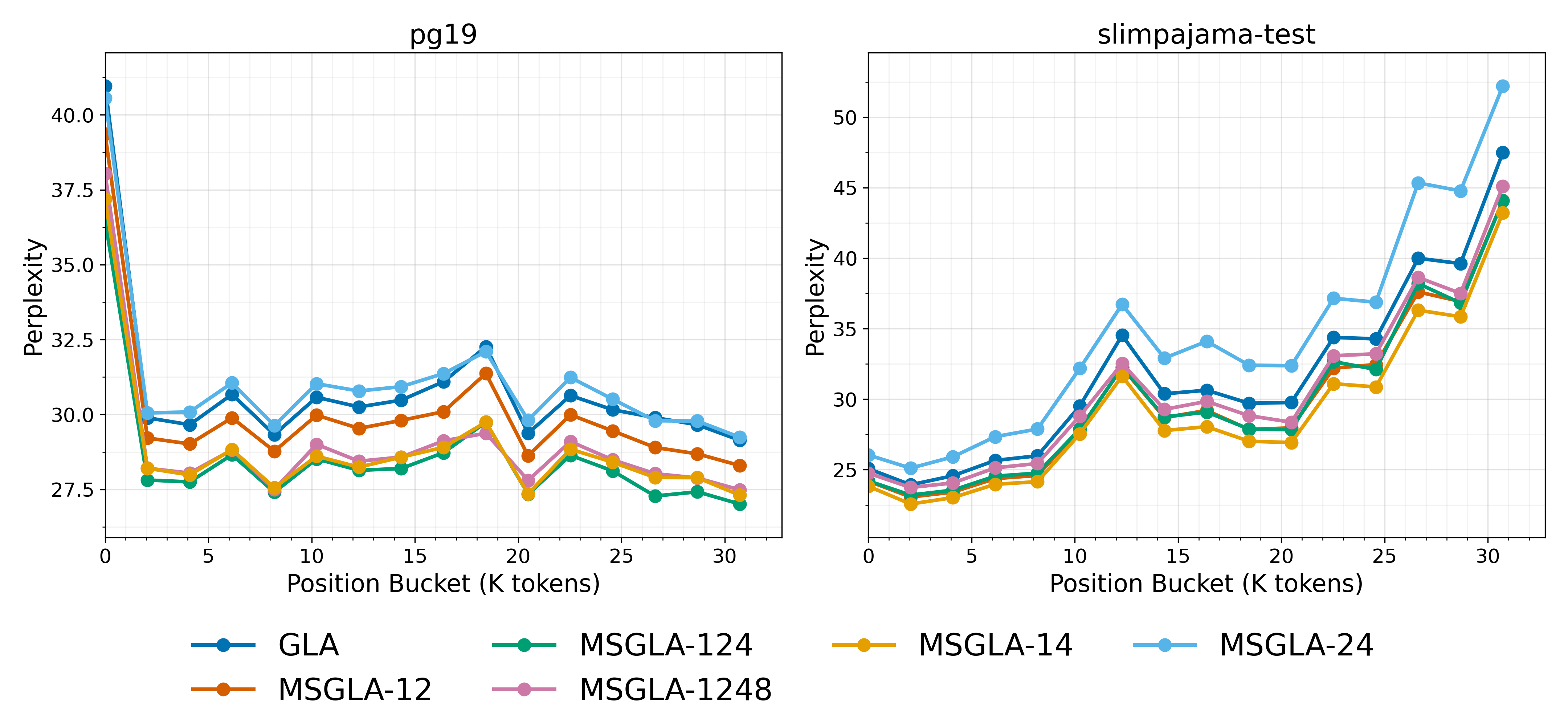}
\end{center}
\caption{Perplexity vs. Position-bucket on two datasets, PG19 (left) and SlimPajama (right) illustrating long-context generalization up to 30,720 tokens.}
\label{fig:perplexity}
\end{figure}

Figure~\ref{fig:perplexity} shows perplexity vs. position bucket for sequences extending up to $15\times$ the training context length, demonstrating the model's ability to handle sequences much longer than those seen during training. While GLA exhibits a persistent upward drift in perplexity beyond the mid-range token positions on both datasets, consistent with a recurrent state that progressively saturates under long-range context. All configurations that include a fine-resolution branch remain below the GLA baseline throughout both datasets. MS-GLA $\{1,2,4\}$ and $\{1,2,4,8\}$ show the flattest profiles overall, sustaining their advantage at the furthest token positions with considerably less degradation than the baseline. On SlimPajama, the characteristic perplexity spike at the mid-sequence document boundary is substantially reduced for the better MS-GLA configurations, suggesting that coarser branches help smooth over abrupt distributional shifts in heterogeneous corpora.

Conversely, the strictly coarse-scale configuration $\{2,4\}$ diverges most severely of all models on both datasets, exhibiting a much steeper degradation than the GLA baseline beyond moderate sequence lengths. This confirms that temporal scale diversity alone does not confer long-context robustness; the fine-resolution branch is a critical architectural requirement.

\subsection{Fusion Layer Routing Stability}
\label{sec:routing-stability}
To verify that the long-context robustness observed above reflects consistent multi-scale
usage rather than the fusion layer collapsing onto a single branch far outside the training
horizon, we measured average fusion routing weights over 32K-token evaluation sequences,
comparing the first, middle, and final 10\% of each sequence. Results are shown in
Table~\ref{tab:routing-stability}.

\begin{table}[t]
\begin{center}
\resizebox{\textwidth}{!}{%
\begin{tabular}{lccccc}
\toprule
\textbf{Region} & \textbf{MS-GLA $\{1,2\}$} & \textbf{MS-GLA $\{1,4\}$} & \textbf{MS-GLA $\{2,4\}$} & \textbf{MS-GLA $\{1,2,4\}$} & \textbf{MS-GLA $\{1,2,4,8\}$} \\
\midrule
First 10\%  & [72.93, 27.07] & [69.89, 30.11] & [49.81, 50.19] & [52.64, 18.41, 28.95] & [47.98, 18.86, 12.33, 20.83] \\
Middle 10\% & [73.45, 26.55] & [70.00, 29.99] & [49.40, 50.60] & [52.91, 18.34, 28.76] & [48.35, 18.58, 12.16, 20.92] \\
Last 10\%   & [72.96, 27.03] & [69.47, 30.53] & [49.25, 50.75] & [52.57, 18.41, 29.03] & [48.04, 18.43, 12.53, 21.00] \\
\bottomrule
\end{tabular}%
}
\end{center}
\caption{Average fusion routing weights (\%) by scale, ordered finest to coarsest, at
different positions within 32K-token evaluation sequences -- roughly $15\times$ the training
context.}
\label{tab:routing-stability}
\end{table}

Routing weights shift only trivially between the beginning and end of these sequences; for
example, in MS-GLA $\{1,2,4\}$ the fine-to-coarse split moves from $[52.64\%, 18.41\%,
28.95\%]$ in the first 10\% to $[52.57\%, 18.41\%, 29.03\%]$ in the final 10\%. This holds
across all variants, confirming that the fusion layer does not collapse onto a single scale
or become unstable well beyond the training horizon. Notably, the model consistently assigns
the majority of its routing weight to the native, finest-resolution branch, reinforcing that
this pathway is the dominant contributor to long-context robustness, rather than an artifact
of reduced per-step capacity in the coarse-only $\{2,4\}$ configuration.

\subsection{Computational Efficiency}
\label{sec:efficiency}
Since our central architectural claim rests on preserving GLA's hardware-efficient
chunkwise training, we report measured TFLOPs, throughput, memory, and wall-clock training cost
for all variants in Table~\ref{tab:efficiency}.

\begin{table}[t]
\begin{center}
\begin{tabular}{lcccc}
\toprule
\textbf{Model} & \textbf{TFLOPs} & \textbf{Throughput} & \textbf{Max mem.\ (GiB)} & \textbf{Train time} \\
\midrule
GLA                  & 91.2 & 31{,}989 & 70.2 & 3d 1:49:25 \\
MS-GLA $\{1,2\}$     & 78.9 & 27{,}619 & 86.7 & 2d 18:23:08 \\
MS-GLA $\{1,2,4\}$   & 84.0 & 29{,}437 & 80.4 & 2d 22:48:41 \\
MS-GLA $\{1,2,4,8\}$ & 77.8 & 27{,}201 & 89.2 & 2d 23:43:25 \\
MS-GLA $\{2,4\}$     & 97.4 & 34{,}123 & 71.5 & 2d 9:11:04 \\
\bottomrule
\end{tabular}
\end{center}
\caption{Training TFLOPs, throughput in tokens/sec, memory, and wall-clock time across GLA and MS-GLA
variants, measured on the training runs described in Section~\ref{train_deets}.}
\label{tab:efficiency}
\end{table}

Coarser branches process proportionally shorter pooled sequences ($\lceil L/s \rceil$
tokens), which offsets much of the overhead introduced by additional branch projections,
pooling, causal upsampling, and fusion, consistent with the structural argument in
Section~\ref{sec:chunkwise}. MS-GLA $\{1,2,4\}$, our best-performing configuration,
retains roughly 92\% of GLA's throughput at only $\sim$14\% higher active memory, while
the purely coarse $\{2,4\}$ configuration exceeds baseline throughput outright since its
branches never operate at full sequence length. Taken together with the accuracy and
robustness gains reported above, these numbers indicate that MS-GLA's improvements come
at a modest, structurally bounded cost rather than through brute-force scaling of compute
or memory.

\section{Discussion}
Across all three evaluation settings, MS-GLA $\{1,2,4\}$ consistently achieves the best overall results, suggesting that three scales including the finest resolution strike the optimal balance between multi-resolution coverage and per-branch representational capacity.

\subsection{The Necessity of Fine-Resolution Branches}
\label{sec:The Necessity of Fine-Resolution Branches}
The most consistent finding across language modeling, recall tasks, and long-context generalization is the decisive role of the finest resolution branch ($s=1$). All configurations that retain at least one branch at native token resolution outperform the GLA baseline in aggregate, while MS-GLA $\{2,4\}$, which omits this branch entirely, degrades across every evaluation setting. Its LAMBADA perplexity collapses to 31.47, its recall F1 falls below the GLA baseline on SWDE and SQuAD, and its long-context perplexity diverges most severely beyond 10K tokens. This pattern strongly suggests that local syntactic processing at the token level is not merely complementary to coarser temporal modeling but is essential i.e. coarse branches alone cannot compensate for the absence of a fine-grained pathway. The long-context results make this particularly clear: without a fine-resolution branch, the model loses the ability to maintain coherent token-level predictions when forced to rely entirely on coarse temporal summaries beyond its training horizon.

\subsection{Scale Coverage vs.\ Branch Capacity}
\label{sec:sc vs bc}
The comparison between $\{1,2,4\}$ and $\{1,2,4,8\}$ illustrates a trade-off between scale coverage and per-branch representational capacity. Adding the coarsest scale ($s=8$) provides marginal or no benefit on language modeling and recall tasks, and the four-scale configuration regresses relative to the three-scale model on both SWDE and FDA (see Table~\ref{tab:recall-results}). This is consistent with head-budget dilution: spreading a fixed head allocation across four scales reduces each branch's representational capacity below the threshold needed for reliable local retrieval. The $\{1,2,4,8\}$ configuration does remain competitive on long-context tasks, however, where the additional coarse scale may provide modest complementary benefit for very distant dependencies.

MS-GLA $\{1,4\}$ presents a complementary case: skipping the intermediate scale ($s=2$) yields the strongest performance on Wikitext perplexity and HellaSwag while lagging on Lambada and average perplexity, suggesting that the resolution gap hurts short-to-medium range coherence even as it benefits broader structural modeling. The one benchmark where the fine-resolution advantage disappears entirely is Winogrande, where GLA and MS-GLA $\{2,4\}$ score competitively with the better configurations. This is consistent with the nature of the task: Winogrande probes short-range commonsense coreference that is less sensitive to long-range temporal structure, and therefore unlikely to benefit from multi-scale decomposition regardless of which scales are included.

\subsection{Long-Context Behavior and Temporal Scale}
The long-context results offer the clearest window into how distributing recurrent
computation across multiple temporal scales affects generalization beyond the training
horizon. GLA exhibits a persistent upward drift beyond 5K tokens on PG19, consistent with a
recurrent state that progressively saturates as context accumulates. MS-GLA $\{1,2,4\}$ and
$\{1,2,4,8\}$, by contrast, maintain notably flatter profiles, suggesting that distributing
computation across scales prevents long-range compression from bottlenecking into a single
resolution. On SlimPajama, the characteristic perplexity spike near the 12--13K token
document boundary is substantially reduced for the better MS-GLA configurations, indicating
that coarser branches help smooth over abrupt distributional shifts in heterogeneous corpora.
This benefit is contingent on retaining a fine-resolution branch, however: coarse branches
alone not only fail to improve long-context generalization but actively degrade
out-of-distribution performance relative to GLA. This is corroborated by the routing
analysis in Section~\ref{sec:routing-stability}, which shows the learned fusion router
consistently favoring the finest-resolution branch even at $15\times$ the training context,
rather than drifting or collapsing under distribution shift.

\section{Future Work}
While our current evaluations establish the efficacy of MS-GLA at the 340M parameter scale,
an essential next step is evaluating its scaling behavior at multi-billion parameter regimes
and over extended training horizons. Beyond scale, MS-GLA opens several promising directions.
Rather than relying on fixed temporal pooling, future work could explore dynamic semantic
resolutions, where the model autonomously aggregates tokens based on linguistic constituents
or informational density instead of fixed positional windows. Temporal resolution could also
be allowed to vary \emph{across} layers rather than sharing one scale set network-wide, letting
lower layers specialize toward local syntax and higher layers toward semantic structure.
Finally, since MS-GLA is a drop-in replacement for standard GLA layers, it is directly
compatible with hybrid architectures that interleave linear and softmax attention, feeding
pooled multi-scale tokens to periodic softmax layers as a natural extension.

\section{Conclusion}
While data-dependent gating has significantly advanced linear attention, our work exposes and resolves a critical remaining limitation: the single-resolution memory bottleneck. We introduced Multi-Scale Gated Linear Attention (MS-GLA) to demonstrate that decoupling recurrent state updates across diverse temporal scales as a potentially effective, solution. By yielding consistent improvements in language modeling, recall accuracy, and long-context extrapolation over the GLA baseline, our findings establish multi-resolution decomposition as a vital architectural bias for linear sequence models. Ultimately, MS-GLA successfully bridges the gap between the expressive power of multi-scale representation and the hardware efficiency of chunkwise training, offering a robust foundation for scaling efficient transformers.

\bibliography{colm2026_conference}
\bibliographystyle{colm2026_conference}

\appendix

\section{Limitations}
\label{sec:limitations}
We note several limitations of this work, most arising from compute constraints (a single
GPU, further bottlenecked by hardware issues during the research window; full hardware
details in Appendix~\ref{app:training_details}).

\textbf{Model scale.} All experiments are conducted at the 340M-parameter, 7B-token scale.
While MS-GLA's added operations (pooling, upsampling, fusion) scale linearly with sequence
length and we have no architectural reason to expect divergent behavior at larger scales,
this has not been empirically validated beyond 340M parameters.

\textbf{Baseline scope.} Our comparisons are scoped to the GLA family rather than including
freshly trained RetNet, Mamba, or Transformer baselines under our exact setup. We rely on
the original GLA paper's comparisons against these architectures, and note that MS-GLA's
consistent gains over GLA should transitively suggest competitiveness with these broader
baselines, though this has not been directly measured.

\textbf{Single seed.} Due to compute constraints, all results are reported from a single
training run per configuration (seed 42; see Appendix~\ref{app:training_details}), without
multiple seeds or significance testing.

\textbf{Ablation coverage.} We ablate the number and choice of temporal scales extensively,
but did not ablate the pooling operator (e.g., average vs.\ max vs.\ last-token pooling) or
alternative fusion mechanisms (e.g., uniform averaging in place of the learned softmax
router), which we leave to future work.

\section{Training Details}
\label{app:training_details}
This appendix provides the full training configuration used for all models (GLA baseline
and MS-GLA variants) reported in the paper, supplementing the summary in
Section~\ref{train_deets}. Exact launch commands are included in the released
\texttt{train.sh} script in our repository.

\paragraph{Optimization.} We use the AdamW optimizer \citep{article-adam} with
$\epsilon = 1\times10^{-15}$, weight decay $0.1$, and a peak learning rate of
$3\times10^{-4}$. The learning rate follows a cosine decay schedule with a minimum
learning-rate ratio of $0.1$, preceded by $3{,}400$ warmup steps. Gradient clipping is
applied at a maximum norm of $1.0$, and non-finite (NaN/Inf) gradient steps are skipped.

\paragraph{Data and batching.} All models are trained from scratch on the FineWeb-Edu
corpus \citep{lozhkov2024fineweb-edu}, tokenized with the tokenizer from
\texttt{fla-hub/transformer-1.3B-100B}. We use a sequence length of $2{,}048$ tokens, a
batch size of $32$, and a gradient accumulation of $1$ step. All models are trained for $106{,}813$ steps to match the
7B-token budget reported in Section~\ref{train_deets}.

\paragraph{Reproducibility.} All runs use a fixed random seed of $42$. Model checkpoints
are saved every $8{,}096$ steps, and training metrics are logged every $10$ steps.

\paragraph{Hardware and parallelism.} All models were trained on a single NVIDIA RTX PRO
6000 Max-Q GPU (\texttt{NGPU=1}, \texttt{NNODE=1}), part of a workstation equipped with an
AMD Ryzen Threadripper PRO 7975WX 32-core CPU and 512GB of RAM; tensor parallelism and
loss parallelism are disabled (\texttt{tensor\_parallel\_degree=1}), and models are
compiled prior to training. The workstation's power supply proved inadequate over parts
of the research window, which constrained the overall compute budget available for this
project and informed the scoping decisions discussed in Appendix~\ref{sec:limitations}. A
single 340M-parameter training run required approximately 2--3 days depending on
configuration (Table~\ref{tab:efficiency}).

\paragraph{Framework.} All models are implemented and trained using the
\texttt{flash-linear-attention} library \citep{yang2024fla} and the \texttt{flame}
training framework \citep{yang2025flame}, built on \texttt{torchtitan}.

\paragraph{Code.} A cleanly documented, modular implementation -- including
scale-specific GLA branches, causal pooling/upsampling utilities, the fusion layer, and
the exact training launch script (\texttt{train.sh}) with per-model configuration files
-- is released at \url{https://github.com/prasoondev/msgla}.
\end{document}